\documentclass[sigconf]{acmart}
\usepackage{soul}
\usepackage[normalem]{ulem}
\AtBeginDocument{%
  \providecommand\BibTeX{{%
    \normalfont B\kern-0.5em{\scshape i\kern-0.25em b}\kern-0.8em\TeX}}}

\copyrightyear{2021} 
\acmYear{2021} 
\setcopyright{acmlicensed}\acmConference[CHI '21 Extended Abstracts]{CHI Conference on Human Factors in Computing Systems Extended Abstracts}{May 8--13, 2021}{Yokohama, Japan}
\acmBooktitle{CHI Conference on Human Factors in Computing Systems Extended Abstracts (CHI '21 Extended Abstracts), May 8--13, 2021, Yokohama, Japan}
\acmPrice{15.00}
\acmDOI{10.1145/3411763.3451591}
\acmISBN{978-1-4503-8095-9/21/05}

\acmSubmissionID{1122}

\begin{document}

% \title{When Will This Autopilot Disengage Itself? A Framework for Autopilot Behavior Understanding}
\title{AutoPreview: A Framework for Autopilot Behavior Understanding}


\author{Yuan Shen}
\email{yshen47@illinois.edu}
\affiliation{%
  \institution{University of Illinois at Urbana-Champaign}
  \city{Champaign}
  \state{Illinois}
  \country{USA}
  \postcode{61820}
}

\author{Niviru Wijayaratne}
\email{nnw2@illinois.edu}
\affiliation{%
  \institution{University of Illinois at Urbana-Champaign}
  \city{Champaign}
  \state{Illinois}
  \country{USA}
  \postcode{61820}
}

\author{Peter Du}
\email{peterdu2@illinois.edu}
\affiliation{%
  \institution{University of Illinois at Urbana-Champaign}
  \city{Champaign}
  \state{Illinois}
  \country{USA}
  \postcode{61820}
}

\author{Shanduojiao Jiang}
\email{sj10@illinois.edu}
\affiliation{%
  \institution{University of Illinois at Urbana-Champaign}
  \city{Champaign}
  \state{Illinois}
  \country{USA}
  \postcode{61820}
}

\author{Katherine Driggs-Campbell}
\email{krdc@illinois.edu}
\affiliation{%
  \institution{University of Illinois at Urbana-Champaign}
  \city{Champaign}
  \state{Illinois}
  \country{USA}
  \postcode{61820}
}

\newcommand{\etal}{\textit{et al}. }
\begin{abstract}
The behavior of self-driving cars may differ from people's expectations (e.g. an autopilot may unexpectedly relinquish control). This expectation mismatch can cause potential and existing users to distrust self-driving technology and can increase the likelihood of accidents. We propose a simple but effective framework, \textit{AutoPreview}, to enable consumers to preview a target autopilot's potential actions in the real-world  driving context before deployment. For a given target autopilot, we design a delegate policy that replicates the target autopilot behavior with explainable action representations, which can then be queried online for comparison and to build an accurate mental model. To demonstrate its practicality, we present a prototype of \textit{AutoPreview} integrated with the CARLA simulator along with two potential use cases of the framework. We conduct a pilot study to investigate whether or not \textit{AutoPreview} provides deeper understanding about autopilot behavior when experiencing a new autopilot policy for the first time. Our results suggest that the \textit{AutoPreview} method helps users understand autopilot behavior in terms of driving style comprehension, deployment preference, and exact action timing prediction. 
\end{abstract}

\begin{CCSXML}
<ccs2012>
   <concept>
       <concept_id>10003120.10003121</concept_id>
       <concept_desc>Human-centered computing~Human computer interaction (HCI)</concept_desc>
       <concept_significance>500</concept_significance>
       </concept>
   <concept>
       <concept_id>10010147.10010178</concept_id>
       <concept_desc>Computing methodologies~Artificial intelligence</concept_desc>
       <concept_significance>500</concept_significance>
       </concept>
 </ccs2012>
\end{CCSXML}

\ccsdesc[500]{Human-centered computing~Human computer interaction (HCI)}
\ccsdesc[500]{Computing methodologies~Artificial intelligence}

\keywords{Autonomous Vehicle, Mental Model, Human Robot Interaction, Imitation Learning, Agent Behavior Understanding, Preview}

\newcommand{\peter}[1]{\textcolor{orange}{#1}}
\newcommand{\niv}[1]{\textcolor{red}{#1}}
%   \includegraphics[width=\textwidth]{graphics/teaser.png}
%   \caption{The frequent major and minor AI model performance updates influence people's mental models about robots' capabilities and intentions in various scenarios. }

\maketitle

\section{Introduction}\label{sec:intro}
Despite recent efforts towards fully autonomous vehicles (e.g., SAE Level 5~\cite{sae_level}), existing self-driving solutions still require human drivers to maintain situational awareness and be ready to take over control at any given moment~\cite{10.1145/3239092.3239559}. 
The effectiveness of these systems requires that the human drivers have developed a reasonable understanding of the autopilot's behaviors and tendencies. 

However, industry currently does not provide sufficient tools to help drivers calibrate appropriate mental models of autonomous technology. We conducted a simple initial survey study to understand how potential and existing users currently explore and build an understanding of autopilot behavior. Our results showed that 59.1\% of our participants expressed that they rarely, if ever, check the content of release notes (the current industry practice), while 77.3\% of our participants indicated they would prefer a previewing tool prior to purchase or deployment. As a result of poor mental model calibration tools, drivers may experience unexpected behaviors when on the road and therefore disengage the autopilot~\cite{why_disable_autopilot}. For example, researchers have found 10.5 hours of YouTube videos that record how autopilot has surprised drivers~\cite{autopilot_trouble}. Our objective is to develop a tool to help drivers become familiar with autopilot behavior, improve their understanding, and establish appropriate levels of trust.

We propose a framework, called \textit{AutoPreview}, which aims to help new or already existing users of autonomous vehicles preview autopilot behaviors of updated control policies prior to purchase or deployment. 
At a high level, \textit{AutoPreview} takes advantage of a delegate model to inform drivers about the potential actions that a target autopilot would take if it were deployed. We implemented a framework prototype in the CARLA simulation environment~\cite{dosovitskiy2017carla}. Our preliminary finding suggests that \textit{AutoPreview} is easy-to-use and can help users better understand autopilot behavior in terms of driving style comprehension, deployment preference, and exact action timing prediction.

\section{Related Work \label{sec:related work}}

Prior work has conducted several studies on building mental models of intelligent agents. These methods can be categorized into online interaction and offline introspection. For online interaction, explainable AI related systems are widely discussed and used~\cite{doi:10.1177/0018720817747730, 8956390, 10.1145/3173574.3174136, kim2018textual}. Through visual or verbal explanations, real-time interaction can directly respond to real-world scenarios but cannot protect users from the danger of unexpected agent behaviors when users have not established sufficient understanding of the agent's policies. As for offline introspection, researchers indicate that end-users can build better mental models of reinforcement learning agent policies either through checking the extracted critical states from the agent trajectories~\cite{huang2018establishing, sequeira2020interestingness, 10.5555/3237383.3237869}, or through actively querying trajectories which satisfy certain behavioral related conditions ~\cite{zhou2020rocus, booth2020bayesprobe, 10.5555/2900929.2900946}. These offline methods offer targeted feedback to users' queries, but require extra effort to explore and thus oppose the principle of least effort~\cite{zipf2016human}. Our method combines the best of both worlds by enabling users to safely and conveniently preview an agent's policies online through real world interaction, via a delegate policy. 

Aside from methods for mental model development, researchers have also explored factors that influence the acceptability of autonomous vehicles. Choi \etal revealed that trust and perceived usefulness strongly affect the user's desire to use autonomous vehicles~\cite{choi2015investigating}. In particular, trust is a widely adopted metric for the level of acceptance of autonomous systems ~\cite{doi:10.1177/0018720819872672, HU201648,doi:10.1177/0018720815625744, ZHANG2019207, choi2015investigating}. They also suggest that system transparency, technical competence, and situation management can positively impact trust, and can therefore indirectly influence the adoption and acceptance of autonomous vehicle.

\section{\textit{AutoPreview} Framework \label{sec:framework}}
The motivation of \textit{AutoPreview} is to make autopilots transparent and understandable to new or active users with no domain knowledge. We aim to provide an easy-to-use and safe tool for these consumers to understand, evaluate, and compare autopilot models before use. Our framework design was guided by the following three design considerations:
\begin{enumerate}
    \item \textit{Safety}: While drivers are learning autopilot behaviors, we should not put them in dangerous situations that may arise as a result of inexperience with the new autopilot system.
    \item \textit{Convenience}: We avoid solutions that require humans to spend extra time reading or learning, a downside of offline introspection as discussed in Section \ref{sec:related work}.
    \item \textit{Realism}: We prefer solutions that enable users to learn autopilot behavior through experiences that are as real as possible. Past work in social psychology has provided strong evidence supporting the fact that realistic experiences yield more clear and accurate attitudes than those developed through unrealistic experiences. ~\cite{fazio1981direct, olson2003attitudes}.
\end{enumerate}

\begin{figure}
\centerline{  \includegraphics[width=0.5\textwidth]{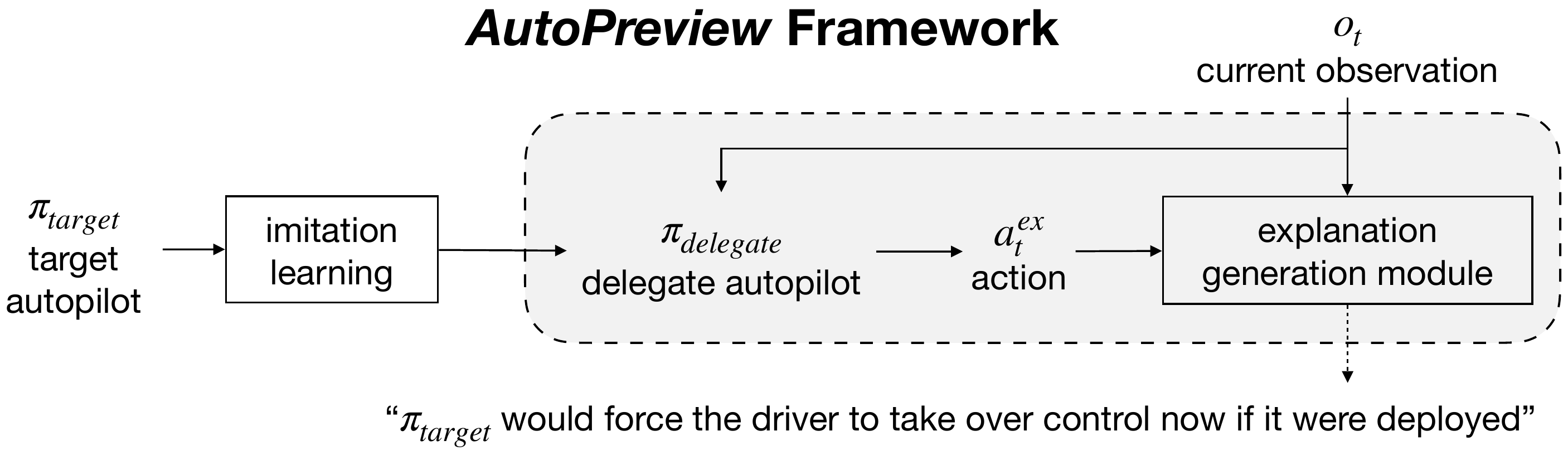}}
\caption{\textit{AutoPreview} Framework. The shaded area represents the framework logic happening inside the end-user's vehicle, and the rest of the logic happens at the self-driving company. The target autopilot, $\pi_{target}$, refers to the model that people are interested in for its behavior. For example, the target autopilot can be a newly released but unfamiliar autopilot that end-users are not sure whether they should deploy. The delegate autopilot, $\pi_{delegate}$, is a delegate model which behavior matches with $\pi_{target}$. The function of $\pi_{delegate}$ is to inform human drivers about the potential actions of $\pi_{target}$ based on the current driving state. However, the proposed actions of $\pi_{delegate}$ will not be executed to control the vehicle.
}
\label{fig:framework}
\end{figure}

Our \textit{AutoPreview} framework achieves the above three design criteria by previewing the behaviors of the target autopilot, $\pi_{target}$, indirectly through a delegate autopilot, $\pi_{delegate}$ (Figure \ref{fig:framework}). The delegate autopilot is generated by imitation learning algorithms and can output control actions that match the behavior of $\pi_{target}$. To clarify, we define a policy, $\pi$, as a function that outputs an action based on an observation. In order to satisfy the safety criterion, a human driver must maintain full control of the vehicle during use of our framework, therefore the control action produced by $\pi_{delegate}$ will not actually be executed. The delegate autopilot will solely inform human drivers about the potential actions of $\pi_{target}$ based on the current driving state. Under the \textit{AutoPreview} framework, drivers can manually control their vehicle to actively learn from interesting scenarios and can evaluate the target autopilot action under those conditions~\cite{felder2009active}. 
We describe the details of our framework in the next subsection. 

\subsection{Details}
Our goal is to enable potential or existing users to preview an autopilot model before purchase or deployment. The initial step in the use of this framework would be to have the desired self-driving car company generate a delegate autopilot model $\pi_{delegate}$, which imitates the autopilot behaviors of $\pi_{target}$. This generated $\pi_{delegate}$ is then delivered to users who are interested in learning about the autopilot behavior of $\pi_{target}$. 

We start by elaborating further on the model training process within the self-driving car company. To achieve the previewing objective while satisfying the previously defined realism criterion, the self-driving car company must send a version of the target autopilot model, $\pi_{target}$, to users so that they can explore the autopilot functionality with online, real-world observations, $o_t$. Note that we consider $\pi_{target}$ as a black-box model with no assumption about its internal structure. Based on this assumption, since the outputs of $\pi_{target}$ are low-level actions (e.g., pedal, brake, steering angle), end users cannot directly map those actions to high-level behaviors (e.g., overtake, change lanes). Our proposed solution is to use imitation learning methods~\cite{kipf2019compositional, kumar2019learning, shiarlis2018taco} to generate a delegate model $\pi_{delegate}$, which matches the behavior of $\pi_{target}$, but remaps the low-level action outputs to the high level action space that humans use to explain and comprehend driving scenarios. \footnote{Note that we assume the delegate accurately captures the target autopilot with sufficient data and computational power. Modeling errors and incorrect abstractions will be explored in future work.}

Once downloaded by the user, the delegate autopilot, $\pi_{delegate}$, can then output high-level actions $a^{ex}_t$ to users based on the current observation, $o_t$. Note that actions $a^{ex}_t$ are what action the target autopilot would take if deployed. As discussed previously, $a^{ex}_t$ will not be executed to control the vehicle. Instead, $a^{ex}_t$ is fed into the explanation generation  module (Figure \ref{fig:framework}) which is responsible for preparing visual or verbal explanation outputs for the user and deciding when to trigger these explanations to avoid counterproductive effects~\cite{10.1145/3313831.3376300, shen2020explain}. 

\subsection{Application}
To demonstrate its practicality, we present two potential use cases of our \textit{AutoPreview} framework, one for existing users, another for potential consumers. \footnote{We also see potential utility of our framework for companies developing self-driving technology (e.g., crowd sourcing early feedback for autopilots in beta release) but leave exploration of this application to future work}
\begin{figure}
\centerline{  \includegraphics[width=0.4\textwidth]{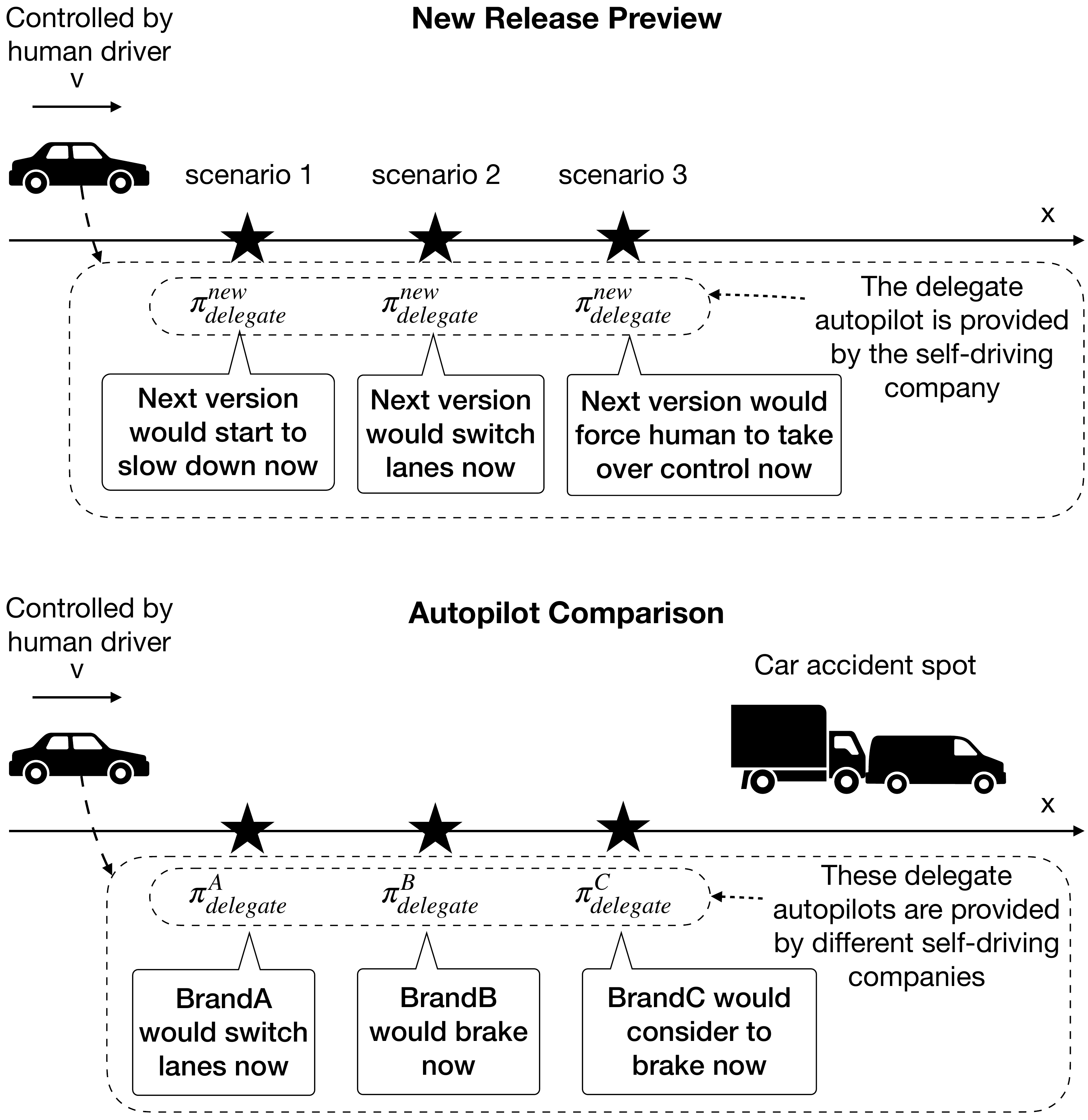}}
\caption{Two potential use cases of our \textit{AutoPreview} framework. The first figure illustrates the previewing feature during an existing self-driving car owner's trial of a new software release involving autopilot changes. The second figure describes the experience a potential buyer of an autonomous vehicle would undergo if using the framework to compare autopilot models in the same driving scenario across brands. BrandA, BrandB, and BrandC represent autopilot models from different companies.}
\label{fig:application}
\end{figure}

\subsubsection{Software Release} As discussed in Section \ref{sec:intro}, existing users of autonomous vehicles need convenient tools to preview autopilot behavior in order to decide if they should deploy a newly released autopilot model. In \textit{New Release Preview} seen in Figure \ref{fig:application}, we illustrate how an existing user could use the framework to safely preview autopilot behaviors when a new software release is available. Directly deploying the newly released autopilot is risky since end-users are unsure about its safety and behaviors. After downloading the delegate autopilot, drivers can manually control their vehicles to actively explore the scenarios they are interested in and evaluate, in those scenarios, the newly released target autopilot's actions based on the output from the delegate autopilot. This previewing feature can enable users to make a deployment decision based on their first-hand experience through our framework. 

\subsubsection{Autopilot Online Comparison} As a second use case, we explore the potential consumers' need to evaluate autopilot performance from different companies prior to making a purchase. Comparing autopilot behaviors across self-driving car companies is a challenging task. Some third-party benchmarking providers have evaluated self-driving cars from different brands based on customized metrics under several test scenarios, but this approach is hard to scale in terms of scenario coverage and car brands. Our delegate autopilot design can compare autopilot behaviors in the same real-world scenarios across different car brands~(\textit{Autopilot Comparison} in Figure \ref{fig:application}). The delegate autopilot has a flexible hardware requirement since it does not need to be trained with the same sensor inputs as the target autopilot~\cite{pan2017agile}. In other words, it is possible to run $\pi_{delegate}$ on different sensor inputs, including non-autonomous vehicles, as long as some sensors are equipped e.g. camera. In \textit{Autopilot Comparison} in Figure \ref{fig:application}, under the same accident scenario, it is easy to tell that autopilot BrandA performs better than the other two brands since its action is the earliest and the most efficient. 

\section{Experiment \label{sec:exp}}
% \raggedbottom
The proposed \textit{AutoPreview} framework enables potential consumers or existing users to preview the behaviors of a target autopilot by observing the actions of a delegate autopilot that shares the same abstracted behaviors as the target autopilot. The goal of this experiment is to investigate what degree of autopilot behavior understanding our \textit{AutoPreview} approach can establish. We conducted a between-subject control experiment online with 10 participants. Our participants are aged between 18 and 30, and agreed to join our study voluntarily. The study took between 30 and 45 minutes for each participant. During the study, we assisted our participants online.  
% In the following subsections, We will introduce the framework prototype, the experiment design, and our preliminary findings .

\subsection{Framework Prototype}
\begin{figure*}[ht]
\centerline{  \includegraphics[width=0.8\textwidth]{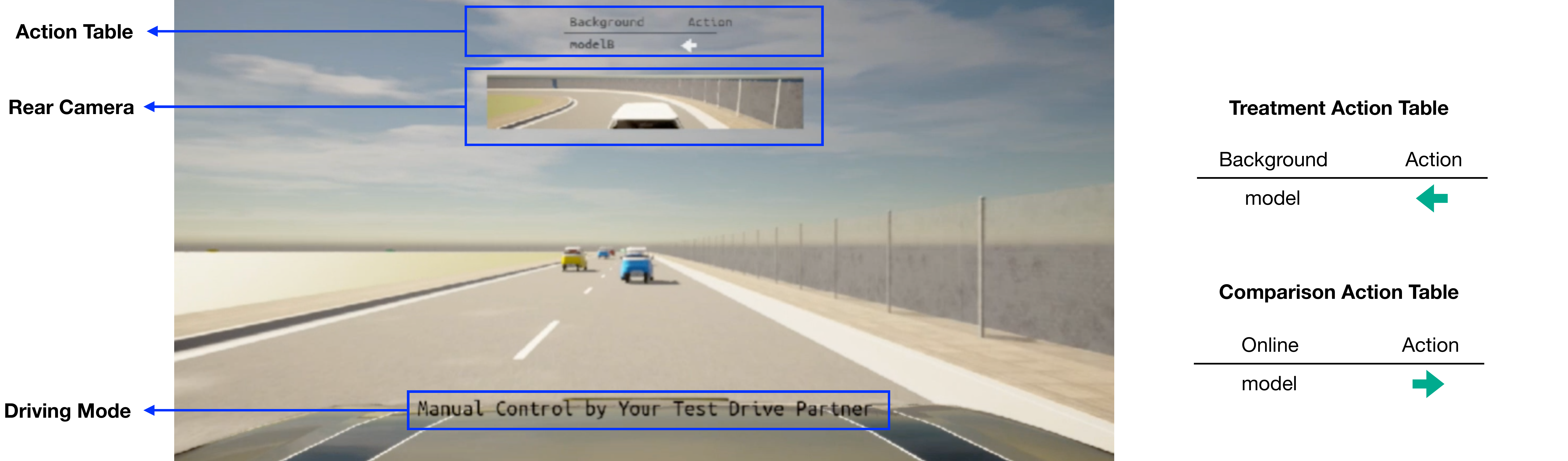}}
\caption{Our framework prototype. The left screenshot shows the driver-perspective video interface that is used in our control experiment. In particular, the action table is different for each experiment group. For the version in the treatment group (upper-right), the action column describes the potential actions based on the delegate autopilot's output. For example, the correct way to interpret the example treatment action table is: the delegate autopilot suggests that the corresponding target autopilot would switch to the left lane if it were deployed to control the vehicle. On the other hand, for the comparison group (lower-right), the arrow icons in the action column reflects the lane-changing operation of the current vehicle which is directly controlled by the target autopilot.
}
\label{fig:carla_interface}
\end{figure*}
We built a prototype of our \textit{AutoPreview} framework in a customized CARLA simulation environment~\cite{dosovitskiy2017carla}. We modified a Model Predictive Control agent provided by CARLA as our target autopilot to control the autopilot behavior and its driving style. The modified autopilot can only perform lane-changing and lane-following operations. Moreover, we explicitly defined the trigger condition of the lane-changing operation. Since we had full access to the target autopilot, the delegate policy's behavior was also defined by explicitly engineered rules.\footnote{Although we did successfully train an imitation learning agent to replicate the MPC behavior as previously described, we did not include the trained agent in our preliminary study since the effect of training error was difficult to control in this initial study.}

Our driving interface was configured to have a first-person viewing perspective together with a rear camera view as shown in Figure \ref{fig:carla_interface}. The actions of the delegate autopilot are presented through an action table. Inspired by the prior work on mental model building via critical states, we considered lane-changing behaviors as critical actions, and only informed participants in real-time about the left or right lane-changing operation of our delegate autopilot through the arrow icons in the action table~\cite{huang2018establishing}. The default maps provided by CARLA contain complicated driving scenarios (e.g., traffic crossings, roundabout, and pedestrian passing), which can be tricky for our participants to observe in a short experiment trial. To reduce the complexity of the driving environment and constrain the set of driving scenarios only to those that include lane-changing or lane-following, we generated a two-lane single-loop map using RoadRunner.

\subsection{Experiment Design \label{sec:exp_design}}
Our experiment hypothesis is that the \textit{AutoPreview} method can help end-users understand the target autopilot's behavior at least as accurate as observing the target autopilot's behavior directly. We measured the degree of autopilot behavior understanding in terms of the aggressiveness level on a 10-point Likert scale. We quantified the degree of understanding in terms of the absolute timing error between the ground truth and user predicted lane-changing timesteps. Specifically, we asked participants to specify the time instance during which the target autopilot would be most likely to switch lanes, along with their level of confidence of this prediction, in eight different five-second test scenarios. We split our participants into two groups, with five participants in each group. In the comparison group, we asked our participants to observe the behaviors of the target autopilot which directly controlled the vehicle. In the treatment group, our participant were told to infer the target autopilot behavior indirectly through the delegate autopilot.
\begin{figure*}[ht]
\centerline{  \includegraphics[width=0.8\textwidth]{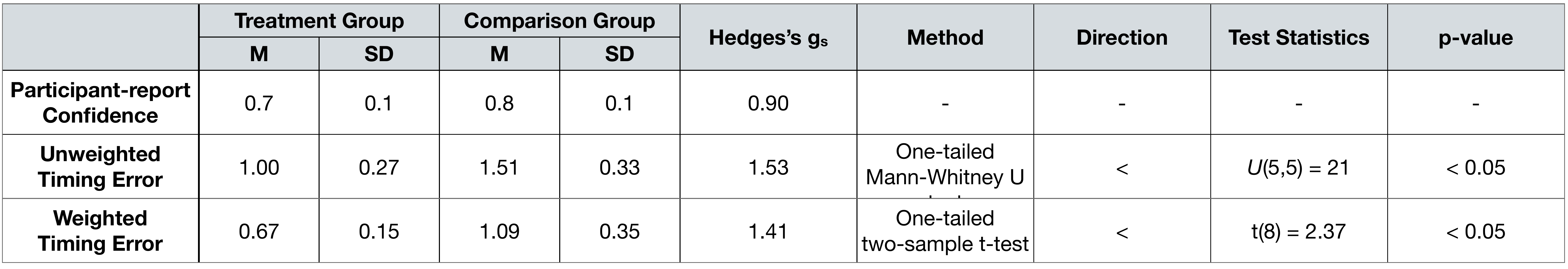}}
\caption{Quantitative result on timing prediction error. Participant-report Confidence refers to how confident participants believe that their predicted lane-changing timing is not 0.5 seconds away from the ground truth. Unweighted Timing Error is the average L1 error between the ground truth and participant predicted lane-changing timing across eight different test scenarios. The unit for the error is seconds. And the Weighted Timing Error is calculated by performing a weighted average of the timing error based on Participant-report Confidence. We calculated Hedges's $g_s$ based on ~\cite{lakens2013calculating}.}
\label{fig:quantitative_result}
\end{figure*}

\begin{figure}
\centerline{  \includegraphics[width=0.5\textwidth]{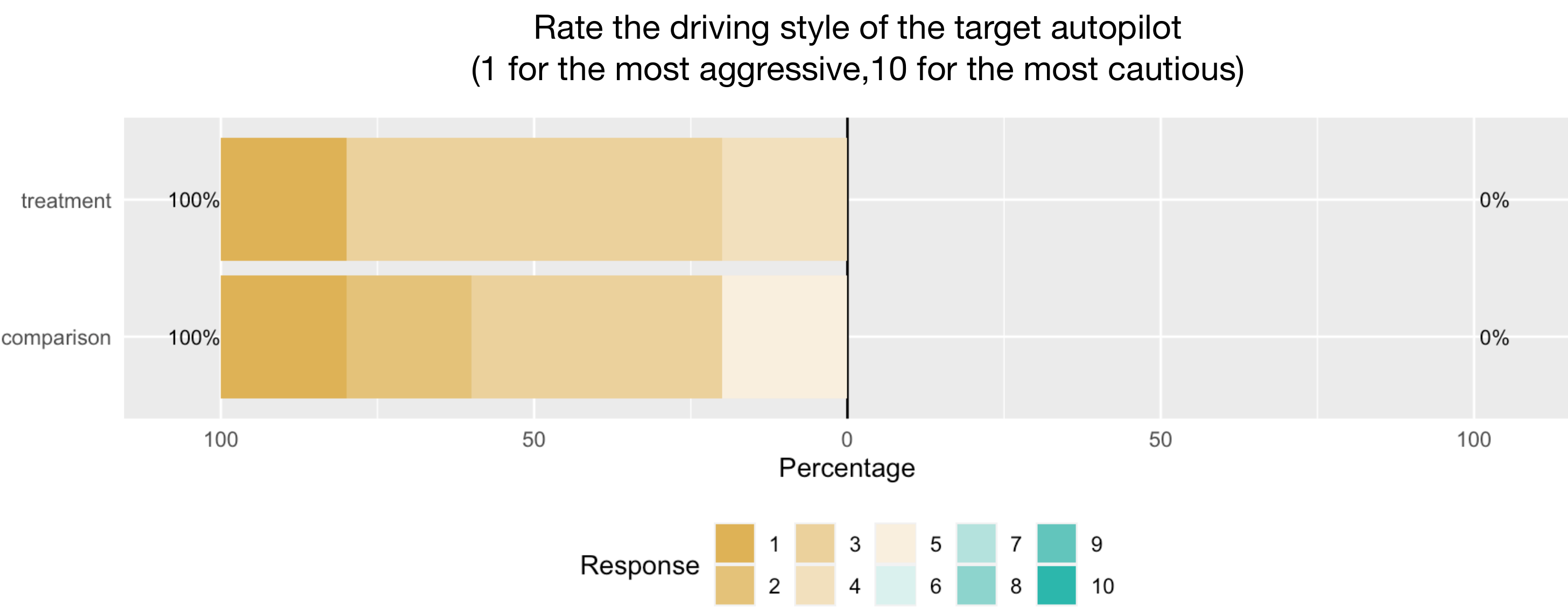}}
\caption{Visualization of participants' overall judgment of target autopilot behaviors for both groups after they learned the target autopilot behaviors. The percent bar plot suggests that participants from both experiment groups agree that the target autopilot has an aggressive driving style. }
\label{fig:qualitative_result}
\end{figure}
To simulate the autopilot experience, we prepared a three minute first-person test-drive video for the different experiment groups (Figure \ref{fig:carla_interface}). Participants in both groups were instructed to imagine themselves as passengers in the car in the video. For the comparison group, the car was directly controlled by the target autopilot. Participants were told that the car was in autopilot mode. As for the treatment group, participants were informed that the car was in manual mode, controlled by a researcher, and the action of the delegate autopilot was presented in the action table. In this preliminary study, we did not enable participants to actively explore the scenarios by interactively controlling the vehicle, in order to reduce experiment noise caused by the participants' exploration strategy. We also ensured consistency in manual mode behavior across different treatment group videos, by using another autopilot to control the vehicle so as to replicate a manually controlled vehicle. Additionally, to reduce the influence of sample bias on our result, we randomly initialized traffic scenarios for each recording such that every video was different. Finally, we explicitly set the target autopilot to have an aggressive lane-changing behavior by controlling the lane switching triggering conditions to ensure a reasonable effect size for our experiment.

The experiment procedure involved three stages: tutorial, virtual test-drive, and post-experiment questions. During the tutorial stage, participants learned about the video interface and their task, and signed the experiment consent form. During the virtual test-drive, the participants imagined themselves as passengers of the car in the video, and finished watching the video without pausing or replaying. While the video was playing, the participants were tasked with figuring out the lane-changing behavior of the target autopilot, based solely on the video content. The post-experiment session then involved an evaluation of the participants' understanding of the target autopilot's lane changing behavior.
% Finally, we evaluated their understanding about target autopilot in the post-experiment session. 

\subsection{Results}\label{sec:exp_results}
We compared the participants' responses from five perspectives: (1) overall autopilot driving style, (2) deployment preference, (3) average action timing error, (4) average prediction confidence, (5) framework usability. As shown in Figure \ref{fig:qualitative_result}, participants in both groups believed the target autopilot had an aggressive driving style with 3 as the majority opinion. As for deployment preference, while 40\% of the participants in the comparison group preferred to deploy, all participants in the treatment group decide not to deploy the target autopilot.

The error and associated confidence of lane change timing (absolute difference between ground truth and user label) is shown in Figure \ref{fig:quantitative_result}. 
We used the Mann-Whitney U test for the Unweighted Timing Error since it does not pass the normality check. Both the weighted and unweighted error show statistically significant difference. 
% To get a sense of how well people can apply the learned knowledge in concrete driving scenarios, we asked our participants to specify their belief about the exact lane-changing moment of the target autopilot in eight different test scenarios. 
% We used the Mann-Whitney U test for the Unweighted Timing Error since it does not pass the normality check. As shown in Figure \ref{fig:quantitative_result}, both the weighted and unweighted error show statistically significant difference. 
Thus, we concluded that the \textit{AutoPreview} method can potentially help potential consumers or target users predict the target autopilot action more accurately than the baseline. Overall, we observed large Hedges's $g_s$. Our explanation is that \textit{AutoPreview} enables users to learn from driving states that are rare in the real world but nonetheless insightful in helping users understand the autopilot's behavior. We believe this advantage can be further leveraged if we enable participants to actively control the vehicle and explore driving scenarios. Finally, for the usability of our framework, in the treatment group, two participants said the delegate autopilot was very easy to use, one said it was easy to use, and one said it was neither easy nor difficult to use in a five-level multiple choice question, leading us to conclude that the framework is, in fact, a viable and convenient solution for the previewing task.
\section{Discussion \& Future Work}
Our preliminary findings suggests that \textit{AutoPreview} can help users intuitively understand autopilot behavior in terms of overall driving style understanding, deployment preference, and exact action timing prediction. 
From our experimental results, we noticed that participants in the treatment group showed less confidence in their timing prediction and more conservative attitude towards deploying the target autopilot model, suggesting that the action table alone is not enough to instill participant confidence in the target autopilot. We consider this a limitation of our framework and attribute the discrepancy in deployment preference between the comparison and treatment groups, as discussed in Section \ref{sec:exp_results}, to this.

There are several limitations with this framework. First, the delegate autopilot can potentially report actions in states that the target autopilot is unlikely to visit, since the delegate autopilot bases action notifications purely on current observations without considering state visitation frequency. Additionally, although the \textit{AutoPreview} framework can protect drivers from the danger of unexpected autopilot behavior during exploration, the notification mechanism we employed might add extra mental load to the driver and can potentially increase the risk of accidents. Furthermore, our prototype can only report information regarding an action triggering moment. Subtle behaviors (e.g., how soft the brake would be), still require further research. As for our experiment, the small sample size as well as the usage of video recordings, ultimately made for a sub-optimal experiment design. We believe a larger sample size as well as the usage of driving simulators or a more interactive tool could potentially yield more conclusive results than those reported in our experiment.

For future work, 
% there are many aspects of this framework to explore further. First, active learning is worth further exploration in our framework. For this project, we used static video as experiment materials. We believe 
we hope to explore whether active learning can improve learning quality. More concretely, we plan to research the improvement in user understanding of autopilot behavior if users are given the ability to control the car and actively create test scenarios that they hope to know autopilot's action in. Furthermore, we hope to explore verbal, textual or augmented-reality-based notification mechanisms in future.

\section{Conclusion}
In this paper, we propose the \textit{AutoPreview} framework, which abstracts autopilot policies into explainable policies for viewing and exploring online. The main contribution of our work is highlighting a novel design space which involves using the preview stage to build or calibrate human drivers' mental model towards the target autopilot. Our preliminary finding suggests that the\textit{AutoPreview} method is easy-to-use and can help users understand autopilot behavior in terms of overall driving style understanding, deployment preference, and exact action timing prediction.
% \section{Acknowledgments}

% Identification of funding sources and other support, and thanks to
% individuals and groups that assisted in the research and the
% preparation of the work should be included in an acknowledgment
% section, which is placed just before the reference section in your
% document.
\newpage
\bibliographystyle{ACM-Reference-Format}
\bibliography{main}
\end{document}